\documentclass[10pt, a4paper]{article}

\usepackage[final]{lrec-coling2024} 

\usepackage{xcolor}
\usepackage{arabtex}
\usepackage{utf8}
\usepackage{comment}
\newcommand{\zaebuc}{ZAEBUC-Spoken}
\newcommand{\hide}[1]{}

\newcommand{\AHAMZAUP}{{\^{A}}}

\newcommand{\AHAMZADN}{{\v{A}}}

\newcommand{\TAMARBUTA}{{$\hbar$}}

\newcommand{\THA}{{$\theta$}}

\newcommand{\SHIN}{{\v{s}}}
\newcommand{\AYN}{{$\varsigma$}}

\newcommand{\AMAQSURA}{{\'{y}}}

\DeclareMathAlphabet{\mathsfit}{T1}{\sfdefault}{\mddefault}{\sldefault}
\renewcommand{\TAMARBUTA}{{$\mathsfit{\hbar}$}}
\renewcommand{\TAMARBUTA}{{$\mathsfit{\hbar}$}}
\renewcommand{\THA}{{$\mathsfit{\theta}$}}
\renewcommand{\AYN}{{$\mathsfit{\varsigma}$}}

\title{{ZAEBUC-Spoken: A Multilingual Multidialectal \\Arabic-English Speech Corpus}}

\name{Injy Hamed,$^{1,2}$ Fadhl Eryani,$^{1,3}$ David Palfreyman,$^{4,5}$ Nizar Habash$^1$} 
\address{
  $^1$New York University Abu Dhabi, $^2$University of Stuttgart,\\
  $^3$University of T{\"u}bingen, $^4$United Arab Emirates University, $^5$Zayed University, Abu Dhabi, UAE\\
  \texttt{\{injy.hamed,fadhl.eryani,nizar.habash\}@nyu.edu}, \texttt{dpalf@uaeu.ac.ae}}

\abstract{
We present {\zaebuc}, a multilingual multidialectal Arabic-English speech corpus. 
The corpus comprises twelve hours of Zoom meetings involving multiple speakers role-playing a work situation where Students brainstorm ideas for a certain topic and then discuss it with an 
Interlocutor. The meetings cover different topics and are divided into phases with different language setups. The corpus presents a challenging set for automatic speech recognition (ASR), including two languages (Arabic and English) with Arabic spoken in multiple variants (Modern Standard Arabic, Gulf Arabic, and Egyptian Arabic) and English used with various accents. Adding to the complexity of the corpus, there is also code-switching between these languages and dialects. As part of our work, we take inspiration from established sets of transcription guidelines to present a set of guidelines handling issues of conversational speech, code-switching and orthography of both languages. We further enrich the corpus with two layers of annotations; (1) dialectness level annotation for the portion of the corpus where mixing occurs between different variants of Arabic, and (2) automatic morphological annotations, including tokenization, lemmatization, and part-of-speech tagging. 
 \\ \newline  }
 
\begin{document}

\maketitleabstract

\setcode{utf8}
\section{Introduction}
Remarkable strides have been recently made in language technologies for distinct, standardized languages. These achievements, however, are not equally met for the vast majority of discourse communities \cite{ranathunga2022some}, including the widespread phenomenon of code-switching \cite{dougruoz2021survey}, which involves the mixing between languages. One main bottleneck that hinders advancements for many languages is the lack of data needed for training NLP models, and more essentially, for evaluation.

In the scope of Arabic, which is a diglossic language \cite{Fer59}, language technologies are usually 
better suited to the formal language, Modern Standard Arabic (MSA), and less proficient 
for regional dialects. On top of the challenges posed by diglossia, Arabic speakers code-switch between MSA and dialects as well as between Arabic and other languages. The former code-switching type is usually limited to formal settings. 
The latter type is prevalent among Arab countries, where code-switching is typically seen between dialectal Arabic and English or French, or both. 

The work presented here is part of the Zayed University Arabic-English Bilingual Undergraduate Corpus (ZAEBUC) Project, which is interested in the study of bilingualism. As an extension to the previously collected ZAEBUC corpus 
\citep{habash2022zaebuc,palfreyman2022bilingual}, 
which focused on bilingual writers, we present a new corpus that is focused on the spoken domain, offering a resource that encapsulates interesting challenges and linguistic phenomena of bilingual speakers. 
ZAEBUC-Spoken corpus is collected through Zoom meetings, with multiple speakers taking part in the conversation. The corpus is multilingual, containing (accented) English, MSA, and two Arabic dialects; Gulf and Egyptian, including speakers from six nationalities. The speakers also code-switch between the four mentioned languages. The corpus includes 
manual transcriptions of the recordings and dialectness level annotations for the portion containing code-switching between Arabic variants, in addition to automatic morphological annotations. 
ZAEBUC-Spoken corpus offers a challenging set to ASR systems given its spontaneous conversational speech nature, as well as an interesting setup to examine the interaction between diverse bilingual speakers.
We make the corpus publicly available.\footnote{\url{http://www.zaebuc.org}\label{fnsite}}

Next, we discuss related work in \S\ref{sec:related_work}. In \S\ref{sec:data_collection} and \S\ref{sec:transcription}, we elaborate on the data collection process and the translation guidelines. 
\S\ref{sec:corpus_stats} presents overall corpus statistics. In \S\ref{sec:cs_stats}, we provide code-switching analyses, including dialectness level statistics.
\S\ref{sec:morhological_analysis} presents the morphological analysis based on the automatic annotations. 
\section{Related Work}
\label{sec:related_work}
In this section we discuss related work with regards to dialectal Arabic speech corpora, Arabic code-switched speech corpora, and morphologically annotated dialectal Arabic  corpora.

With regards to \textbf{dialectal Arabic speech corpora}, 
numerous efforts provided dialectal Arabic resources for speech recognition. 
The MGB-2 Arabic challenge dataset \citep{ABG+16} comprising of 1,200 hours of speech gathered from Aljazeera Arabic TV channel contains a portion of dialectal Arabic. While the majority of the corpus is MSA speech, a subset of the corpus (estimated 30\%) contains dialectal Arabic speech, including Egyptian, Gulf, Levantine, and North African dialects. \citet{almeman2013multi} also collected the Multi Dialect Arabic Speech Parallel Corpora, containing around 32 hours of speech covering MSA as well as Gulf, Egyptian and Levantine dialects. 
Other efforts have targeted specific dialects. 
For Egyptian Arabic, several corpora are available, including CALLHOME Egyptian Arabic corpus \citeplanguageresource{GKA+97} and MBG-3 \citep{AVR17}. 
For Gulf Arabic, the Gulf Arabic Conversational Telephone Speech corpus \citeplanguageresource{Appen:2006:gulfSpeech,Appen:2006:gulfTranscripts} consists of around 46 hours of spontaneous Gulf Arabic speech obtained from telephone conversations. 
\citet{elmahdy2014development} also present the 15-hour QA corpus, collected from TV series and talk show programs. 
Other corpora have covered other dialects including Moroccan Arabic \citep{ali2019mgb} and Levantine Arabic \citeplanguageresource{Appen:2005:BBN_AUB,Appen:2007:levantineTranscripts,maamouri2007fisher}. 

From the perspective of \textbf{code-switched Arabic speech corpora}, the presented corpus offers an interesting setup, having two different scenarios of code-switching produced by Arabic speakers; code-switching between Arabic and foreign languages and code-switching between Arabic variants.  Similar to our corpus, other researchers have also collected code-switched corpora, 
however, the corpora usually focus on either of the two code-switching scenarios. 
\citet{Ism15} gathered 89 minutes of speech containing Saudi Arabic-English code-switching through informal dinner gatherings. 
\citet{AAM18} collected a 7.5 hour corpus containing code-switched Algerian Arabic-French gathered from informal conversations as well as read speech of books and movie transcripts. \citet{HVA20,hamed2022arzenST} collected the ArzEn corpus containing 12 hours of code-switched Egyptian Arabic-English speech through informal interviews, and commissioned  their English translations. 
\citet{chowdhury2021towards} presented the Economic and Social Commission for West Asia (ESCWA) corpus containing 2.8 hours of United Nations meetings. The corpus contains code-switching between Arabic (including different dialects) and English/French. In \citet{mubarak2021qasr}, 2,000 hours of speech were collected from Aljazeera news channel, where 0.4\% of the corpus ($\sim$6,000 utterances) have code-switching between Arabic and English/French.  
In the direction of covering code-switching between dialectal Arabic and MSA with language identification, \citet{chowdhury2020effects} annotate a 2-hour subset from ADI-5 development set in the MGB-3 challenge \citep{AVR17} containing Egyptian Arabic-MSA code-switching for word-level language identification. 

With regards to \textbf{morphologically annotated dialectal Arabic corpora}, the amount of available resources varies across dialects. Egyptian Arabic, receiving significant attention, is supported by several corpora, including the Egyptian Colloquial Arabic Lexicon \citeplanguageresource{Kilany:2002:egyptian} and the Egyptian Arabic Treebank 
(\citealplanguageresource{Maamouri:2012:arz}; \citealp{Maamouri:2014:developing}). 
Less corpora are available for Gulf Arabic. \citet{Khalifa:2016:large,Khalifa:2018:morphologically} collected the Gumar corpus from 1,200 forum novels and extended a subset of the corpus with morphological annotations as well as orthographic modifications following the Conventional Orthography for Dialectal Arabic (CODA) guidelines \cite{Habash:2018:unified}. 
\citet{habash2022zaebuc} also collected the ZAEBUC corpus, an
Arabic-English bilingual writer corpus containing short essays written by first-year university students along with assigned writing proficiency ratings. The corpus was manually corrected for spelling and grammar errors, and annotated for morphological tokens, part-of-speech (POS) tags, and lemmas.  Other corpora exist for other dialects including Jordanian Arabic \citep{Maamouri:2006:developing}, Palestinian Arabic \citep{Jarrar:2016:curras}, as well as Moroccan and Sanaani Yemeni Arabic \citep{Al-Shargi:2016:morphologically}.
In the context of code-switching, previous efforts also targeted collecting code-switched Egyptian Arabic-English corpora with morphological annotations including morphological segmentations, POS tags and lemmatization \citep{BHA+20,gaser2022exploring}. 

In comparison to previously mentioned corpora, our \textbf{contribution} stands out on several dimensions. 
Firstly, the corpus is obtained through Zoom meetings, which opens up possibilities for extending the corpus to other tasks, such as meeting summarization. 
The corpus also 
contains multiple Arabic variants (MSA, Gulf Arabic, and Egyptian Arabic), accented English, as well as code-switching across the languages and dialects. We also provide automatic morphological annotations, including tokenization, lemmatization and POS tagging. Unlike the previously mentioned corpora where such annotations were provided for textual data, spontaneous speech data introduces challenges to POS tagging, 
as the normal flow of sentences may be broken due to repetitions and/or corrections. We currently only provide automatic annotation, which we plan on extending with manual revisions. Moreover, as part of our work, we present our transcription guidelines handling issues arising due to the spontaneous nature of the corpus as well as code-switching.

\section{Data Collection}\label{sec:data_collection}
The recordings were collected through Zoom meetings in which two Students and an Interlocutor simulated 
a work situation relevant to the students' major. Topics included \textit{employee health and wellbeing}, \textit{studying abroad}, \textit{arts and design}, \textit{SWOT analysis}, \textit{advertising}, and \textit{tourism}. 
The Students were asked to prepare ideas to present to an Arabic-speaking or English-speaking Interlocutor of senior status (e.g., manager, dean, etc., depending on the topic). Afterwards, the Interlocutor, a person unknown to the Students, joined them to hear about and discuss their ideas. The interactions were set up by one of the authors, referred to as the Moderator. Each meeting consists of four phases:
\begin{enumerate}
    \item \textbf{\textit{Phase 1}:} The Moderator introduces the task to both Students, showing an email from the Interlocutor requesting their ideas for a specific purpose. This phase is conducted in English, except where Students are asked to read aloud a task which is written in MSA for an Arabic-speaking Interlocutor.
    \item \textbf{\textit{Phase 2:}} The two Students discuss the task together. They are allowed to converse in any language, as they prefer. This phase usually contains a mix of Gulf Arabic and English.
    \item \textbf{\textit{Phase 3}:} The Students present their ideas to the Interlocutor, who stimulates further discussion of the task at hand. There are two options for this phase, where the Interlocutor talks in either English or MSA. In the first case, the Students 
    use the English language; in the second case, the Students are allowed to choose between MSA and dialectal Arabic. In the case of Arabic-speaking Interlocutors, 
    we observe code-switching between Arabic and English in addition to 
    code-switching between MSA and dialects which arises due to the Egyptian Interlocutors primarily speaking in MSA with slight use of Egyptian Arabic and Students using a mixture of MSA and Gulf Arabic. 
    \item \textbf{\textit{Phase 4}:} The Moderator ends the meeting. This phase is conducted in English.
\end{enumerate}
A total of 14 meetings were conducted, with equal distribution among both setups; Arabic-speaking and English-speaking Interlocutors. 
Overall, 16 Students took part in the recordings, 
where each Student participated in a maximum of two meetings; one with an Arabic-speaking and the other with an English-speaking Interlocutor.\footnote{Information about the recordings and participants are released as part of the corpus metadata. }
We also include a 15$^{th}$ recording as part of the corpus, which was a pilot recording collected as part of the development process, only consisting of \textit{Phase~2}. The Zoom meetings are audio recorded, with the audio input of all participants saved in a single audio file. 
We also obtain separate single-channel recordings for 
participants' audio streams, where the audio from each participant is saved as a separate file. This is obtained for 11 out of the 15 recordings, due to technical reasons.
This setup allows researchers working on speech recognition to utilize the corpus as needed with regards to overlapping speech.

\paragraph{Overview of the Participants: }
Across all meetings, there is one English-speaking Moderator. All Students are Emirati and all but one 
are female. 
The Students are enrolled in different majors, with four Students in \textit{International Studies}, three in \textit{Communication and Media}, two in \textit{Finance}, two in \textit{Animation Design} and the remaining Students each in \textit{Psychology}; \textit{Public Health and Nutrition}; \textit{Multimedia Design}; \textit{Interior Design}; \textit{Marketing and Entrepreneurship}; \textit{Human Resources}; and \textit{Accounting}. 
With regards to the Interlocutors, there are two Egyptian Arabic speakers and six English speakers coming from different countries: United Kingdom (3), Greece (1), Austria (1), and China (1), introducing a range of different English accents. 

\section{Transcription}\label{sec:transcription}
\begin{figure*}[t]
\centering
 \frame{\includegraphics[width=\textwidth]{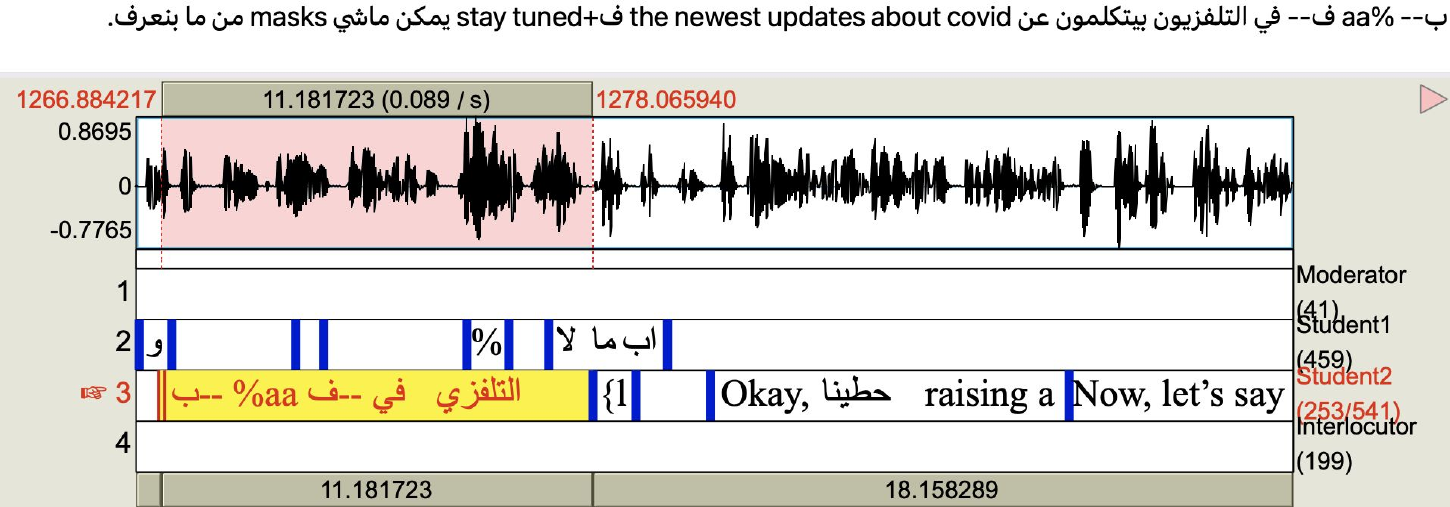}}
   \caption{Example of transcription using Praat.}
  \label{fig:praat}
\end{figure*}

The collected recordings were manually transcribed. 
The transcribers were allowed to choose between using Praat \citeplanguageresource{Praat} or ELAN \citep{brugman2004annotating}. 
In both cases, four tiers were used, 
with each corresponding to one of the four speakers 
(Moderator, two Students, and Interlocutor), as shown in Figure~\ref{fig:praat}. 
The annotation team on the project comprised of three Arabic-English bilingual annotators: \textit{A1} (Emirati Arabic native speaker), \textit{A2} (Egyptian Arabic native speaker), and \textit{A3} (Yemeni Arabic native speaker with extensive residency in the UAE). 

All recordings were initially transcribed by 
annotator \textit{A1}. 
The annotation at this round also involved segmentation of utterances, where an utterance is defined as a segment of speech that is grammatically and semantically complete (except for unfinished or interrupted utterances). The preferable minimum length of an utterance is 10 seconds, and the maximum is 30 seconds unless it is not possible to divide the segment. 
To ensure transcription quality, each recording underwent two rounds of revision performed by annotators \textit{A3} followed by \textit{A2}. 
The final checks demonstrate good quality of transcriptions and minimal errors.
The recordings and transcriptions are made available, in addition to ASR files following Kaldi \cite{PGB+11} data preparation format to further facilitate the use of the corpus for speech recognition purposes.\footref{fnsite}

\subsection{Transcription Guidelines}
In this section, we discuss the transcription guidelines. The guidelines cover four categories; general transcription rules, 
conversational speech transcription rules, 
code-switching transcription rules, 
and orthography rules for English and Arabic. We mostly rely on Callhome \citeplanguageresource{GKA+97} for the general and conversational speech transcription rules, on ArzEn speech corpus \citep{HVA20} for rules related to code-switching, on the SRI Speech-based Collaborative Learning Corpus (SBCLC) 
(\citealp{richey2016sri}; \citealplanguageresource{SBCLC}) 
for rules related to transcribing English words, and on CODA guidelines \cite{Habash:2018:unified} for dialectal Arabic orthography decisions. We mark transcription decisions that are based on Callhome, SBCLC and ArzEn corpora with $H$, $C$ , and $A$, respectively. Afterwards, we discuss CODA guidelines and elaborate on specific decisions that were made within the scope of 
our corpus. 
\subsubsection{General Transcription Rules (GR)}
\paragraph{[\textit{GR-punctuation}]} For punctuation, transcribers are requested to use punctuation as they see fit.

\paragraph{[\textit{GR-numbers}]$^H$} Numbers are written in full text, rather than digits.

\paragraph{[\textit{GR-background}]$^H$} For background noise and typing sounds, the following annotation is used: \textit{[/noise] transcription [noise/]} and \textit{[/typing] transcription [typing/]}, where these tags surround the transcriptions with overlapping background sound. If the sound does not overlap with transcriptions, the annotation is inserted with empty transcription, for example [/noise] [noise/]. We also use the following annotation \textit{[/reading] transcription [reading/]} to denote that a person is reading text aloud. 

\paragraph{[\textit{GR-unclear}]$^H$}
For unclear words, transcribers are requested to listen to the audio multiple times, including making use of the single channels. The transcription of the unclear words is placed between double parentheses, as \textit{((transcription))}. In case the words are unclear due to corruption in the audio file, this is marked as \textit{(([transcription]))}. In case the transcribers are not able to make a guess, the parentheses are placed with empty transcription. 

\paragraph{[\textit{GR-mispronounciation}]$^H$} If a speaker mispronounces a word, the intended word is transcribed, regardless of its pronunciation, and is surrounded by equal sign as \textit{=mispronounced word=}.

\paragraph{[\textit{GR-newTerm}]$^H$} If a speaker comes up with a new term, the term is transcribed between double asterisks as \textit{**term**}.

\subsubsection{Conversational Speech Rules (CR)}
\paragraph{[\textit{CR-repetition}]$^H$} The transcription needs to be an exact reflection of what was said, including repetitions and incomplete words. 

\paragraph{[\textit{CR-partial}]$^H$} Partial or incomplete words are marked with `\texttt{-{}-}', such as \textit{arch\texttt{-{}-}} and \textit{\texttt{-{}-}tective}.

\paragraph{[\textit{CR-flow}]$^A$} Given the spontaneous nature of the corpus, it is common to find disfluencies that break the flow of sentences, including repetitions, corrections, and changing course mid-sentence. We use double dots (..) within text 
to mark such cases as needed to help with the readability of the transcriptions. 
We also use (..) at the end of unfinished utterances when the speaker stops mid-sentence.


\paragraph{[\textit{CR-interruption}]} We mark short interruptions within text with a tilde ($\sim$). We also use ($\sim$) to mark the end of unfinished utterances due to the speaker being interrupted.

\paragraph{[\textit{CR-nonSpeech}]$^H$} For non-speech segments, we use the following tags:
\{laugh\}, \{cough\}, \{sneeze\}, \{breath\}, \{lipsmack\}, \{pause\}, \{gasp\}, \{shush\}, and \{hem\} (clearing throat).

\paragraph{[\textit{CR-interjections}]$^H$} Interjections should be preceded with a percentage sign (\%). 
We use the following closed set of Arabic and English interjections: 
\%oh,\<أوه>\%, \%aa, \<أأ>\%, \%um,  \<ام>\%, \%mm, \<مم>\%, \%hm, \<هم>\%, \%ah, \<آه>\%, \%aha, \<أها>\%, \%ehm, \<إهم>\%, \%nn,\<إن>\%, \%ha, \<ها>\%, \%tt, \<تت>\%, \%er, \<ار>\%, and \%oops. 
In case of extra long interjections (such as `Oooooh'), a colon is added to the interjection, such as `\%oh:'. The choice of script for the interjection is based on the main language of the utterance.

\subsubsection{Code-switching Rules (CSR)}
\paragraph{[\textit{CSR-script}]$^A$}
Arabic words are written in Arabic script and English words 
are written in Latin script. For English words that are commonly used in Arabic (loanwords), the choice of script depends on the pronunciation. If the word is commonly used in Arabic and pronounced in Arabic accent, then Arabic script is used. If commonly used and pronounced in non-Arabic accent, Latin script is used. 

\paragraph{[\textit{CSR-MCS}]$^A$} For morphologically code-switched words, where Arabic affixes and/or clitics are attached to English words, the following annotation is used:
<Arabic Morphemes in Arabic script>+<English word in Latin>+<Arabic Morphemes in Arabic script>, e.g., 
\<وا>+implement+\<ي> 
\textit{y+implement+wA} `they implement'.

\subsubsection{English Orthography Rules (ENR)}
\paragraph{[\textit{ENR-dictionary}]$^C$} American English spelling is used throughout transcriptions, and grammatical errors are not corrected.
\paragraph{[\textit{ENR-contractions}]$^C$}
Standard contractions are used when the contracted pronunciation is used, otherwise, the complete form is used. This also applies to non-standard contractions like `gonna' and `wanna'.
\paragraph{[\textit{ENR-acronyms}]$^C$} If an acronym is pronounced like a word, it is written in uppercase without spaces, such as AIDS. Acronyms pronounced as the individual letters are written in uppercase where letters are separated by underscores, such as U\_A\_E. 

\paragraph{[\textit{ENR-letters}]$^C$}
When the speaker utters single letters, including the case of spelling out a word, the letters are transcribed separately in upper case, for examples:
`We will go for plan B.'

\subsubsection{Arabic Orthography Rules}
As described in Section~\ref{sec:data_collection}, the Arabic component of this corpus comprises MSA, dialectal Arabic, and a rich mix of both. 
As the official language of all Arab countries, MSA enjoys a well-defined official orthographic standard which we follow. For dialectal speech, we follow the Conventional Orthography for Dialectal Arabic (CODA), which is an on-going effort to specify orthographic conventions for a growing number of Arabic dialects. CODA has been used in a number of large-scale Arabic Dialect projects \cite{Habash:2018:unified, Bouamor:2018:madar,Khalifa:2018:morphologically}. 

A core precept in CODA is to spell root radicals using an MSA cognate as a reference, according to a defined list of the most common sound to letter correspondences in Arabic.\footnote{\url{coda.camel-lab.com/\#4321-root-radical-spelling}} This follows from the observation that when writing dialectal Arabic without conventional rules, spelling tends to reflect a tension between spelling according to the phonology of a given utterance on the one hand, and the spelling of a closely related MSA cognate on the other. For instance the Gulf word /w~aa~y~i~d/\footnote{CAPHI phonetic scheme \cite{Habash:2018:unified}.} meaning ‘very’ or ‘a lot’, may be spelled \<وايد> \textit{wAyd}, reflecting its phonology, or \<واجد> \textit{wAjd}, reflecting its MSA cognate.\footnote{HSB transliteration scheme \citep{Habash:2007:arabic-transliteration}.}

CODA regulates this tension by prioritizing the use of MSA cognates as references, more or less familiar to all Arabic speakers. At the same time CODA aims to “strike an optimal balance between maintaining a level of dialectal uniqueness and establishing conventions based on MSA-DA similarities” \cite{Habash:2018:unified}. One way this balance is struck is by allowing specific rules for certain morphemes, often highly marking, such as allowing the pronominal 2nd person feminine clitic when it is pronounced /tsh/ to be spelled with \<ج> \textit{j}, as most Gulf writers prefer, such as in the word /3~i~n~d~i~tsh/ \<عند+ج> \textit{{\AYN}nd+j}, ‘with you~[fs]’. Beyond root radicals, CODA also regularizes templatic pattern spelling according to an MSA reference modulo regular minor sound changes, e.g.: /(2~i)~t.~t.~a~w~w~i~r/ meaning ‘to develop’ is spelled according to how it would be spelled in MSA, \<تطور> \textit{tTwr}, instead of how it is pronounced \<اتطور>* *\textit{AtTwr}.

Efforts employing CODA have thus far dealt with text based data, but this project marks the first time annotators have used it for representing a speech corpus. This has called for additional specifications on two CODA rules.  The first rule involves hamzas (glottal stops) that appear at the beginning of basewords as part of Alif-Hamza letters (i.e., \<أإ>~\textit{{\AHAMZAUP}{\AHAMZADN}}). Whereas previous CODA annotations have stripped baseword initial hamzas (to \<ا> \textit{A}), our annotations spell out these hamzas when they are \textit{audibly pronounced} in the recording.
%
%
The second rule involves the spelling of particles which have alternative proclitic forms with \textit{shortened vowel}, e.g.,
 \<شو> \textit{{\SHIN}w} /sh~uu/ and
 +\<ش> \textit{{\SHIN}+} /sh~u/ `what'.
%
In CODA both forms are valid and depend on vowel length, which may not always be evident. As such we opted to always spell their non-clitic elongated form.



\subsubsection{Redacting Participants' Names}
In the public release, we redact the mentions of participants' names during the meetings, to protect the participants' identities. This is done on the word-level, where the name mentions are bleeped in the audio files and the names in the transcriptions are replaced with the following references: <Moderator>, <Student1>, <Student2>, and <Interlocutor>.

\subsection{Dialectness Level Annotation}
\label{sec:LID_annotation}
As previously mentioned, the corpus contains code-switching between Arabic and English mainly occurring in \textit{Phase 2} as well as between MSA and dialects occurring in \textit{Phase 3}. 
While the former type of code-switching can be automatically detected as Arabic and English use different scripts which is maintained by our transcription guidelines, the distinction between MSA and dialectal words in the latter code-switching type is a challenging task. This is not only due to the shared script, but also due to the shared vocabulary. 
There have been a number of efforts on defining and measuring the degree of dialectness in Arabic \cite{habash2008guidelines,zaidan2011arabic,Keleg:ALDI}. 
In this work, we follow the guidelines introduced in \citet{habash2008guidelines}
 where five levels are defined for annotating Arabic dialectness:
\begin{description}
    \item[\textit{L0}] denotes perfect MSA.
    \item[\textit{L1}] denotes imperfect MSA. This includes utterances with  nonstandard forms, such as syntax or morphology that is inclined towards dialects; however it does not include any strong dialectal markers.
    \item[\textit{L2}] denotes MSA-dialectal code-switching. This includes utterances having strong dialectal markers where the contribution of dialects is nearly equal to or less than MSA.
    \item[\textit{L3}] denotes dialect with MSA incursions. The utterance is mostly dialectal, with some embedded MSA words.
    \item[\textit{L4}] denotes pure dialect.
\end{description}

We perform manual dialectness level annotation on the utterances in \textit{Phase 3} for the seven recordings having Arabic-speaking Interlocutors. We only annotate \textit{Phase 3}, as it is the phase containing MSA-dialectal Arabic code-switching. 
The annotators are asked to listen to each utterance and annotate it according to the guidelines mentioned above. 
The annotation is placed at the start of the utterance transcription followed by `||'. For example, for an utterance identified as \textit{L2}, the annotation is: ``L2|| \textit{transcription}''.
The annotation is performed by annotator \textit{A1} and afterwards revised by annotator \textit{A2}. Cases of disagreements, comprising 22.5\% of utterances, were annotated by annotator \textit{A3}, providing the final decisions for the annotations.

\section{Corpus Statistics}\label{sec:corpus_stats}
\begin{table}[t]
\centering
\begin{tabular}{|l|r|}
    \hline
      \multicolumn{1}{|c|}{\textbf{Category}}&\multicolumn{1}{c|}{\textbf{Value}}\\
      \hline\hline
      \# Speakers & 27\\
      \# Moderator & 1\\
      \# Students & 16+2(pilot)\\
      \# Interlocutors & 8\\
      \hline
      \# Meetings &15\\
      Total Duration (h) & 11.9\\
      Speech Duration (h) & 10.5\\
      Average Meeting Duration (h) & 0.8\\
      \hline
      \# Utterances & 6,033\\
      \# Tokens & 94,101\\\hline
\end{tabular}
\caption{Corpus Size Overview}
\label{table:general_stats}
\end{table}
Table \ref{table:general_stats} presents general corpus statistics, including the number of speakers, the corpus size in hours, and the total number of tokens and utterances. Among  utterances containing language-specific words (not solely annotations),  the average utterance duration is 7.2 seconds containing on average 17.7 tokens. 
Table \ref{table:corpus_stats} presents token-level and utterance-level statistics across recordings with Arabic- and English-speaking Interlocutors. 
On the token-level, we report the number of Arabic, English, and morphologically code-switched (MCS) words and partial words. We also report counts for punctuation, digits, and several annotations. 
On the utterance-level, we report the counts for monolingual Arabic, monolingual English, and code-switched Arabic-English utterances as well as those composed only of annotations. For the code-switched utterances, we also present a breakdown of their counts according to the extent of code-switching, measured as the percentage of English words in the utterance. With regards to \textit{GR-unclear} annotations due to corruption in the audio file, it is found in 152 utterances (2.5\% of the corpus), 
with 174 instances of the annotation.

\begin{table}[t]
\small
\centering
\setlength{\tabcolsep}{4pt}
\begin{tabular}{|l|r|r|r|}
\hline
\multicolumn{1}{|c|}{\textbf{Type}}&\textbf{Rec-Ar}&\textbf{Rec-En}&\multicolumn{1}{|c|}{\textbf{Total}}\\
\hline
\hline
\multicolumn{4}{|c|}{\textbf{Token-level Analysis}}\\
\hline

Arabic words&24,301&3,571&28,515\\
English words&9,692&33,059&43,767\\
MCS words&233&76&346\\
Arabic partial words&932&81&1,015\\
English partial words&71&241&317\\
MCS partial words&4&3&7\\
\hline
Punctuation&5,078&5,741&11,112\\
Digits&0&0&0\\
GR-background&96&126&238\\
GR-unclear&708&826&1,610\\
GR-mispronounciation&118&80&198\\
GR-newTerm &0&0&0\\
CR-flow&598&730&1,366\\
CR-interruption&137&160&318\\
CR-nonSpeech&291&187&500\\
CR-interjections&2,571&2,155&4,792\\
\hline
Total words (full+partial)&35,233&37,031&73,967\\
Total tokens&44,830&47,036&94,101\\
\hline
\hline
\multicolumn{4}{|c|}
{\textbf{Utterance-level Analysis}}\\
\hline
Monolingual Arabic&1,577&200&1,840\\
Monolingual English&584&1,808&2,424\\
Code-switched Ar-En&549&344&999\\
\hspace{0.3cm}English: 1-20\%&274&80&359\\
\hspace{0.3cm}English: 21-40\%&102&54&180\\
\hspace{0.3cm}English: 41-60\%&57&40&121\\
\hspace{0.3cm}English: 61-80\%&38&49&114\\
\hspace{0.3cm}English: 81-99\%&78&121&225\\
Annotations only&504&242&770\\
\hline
Total Utterances&3,214&2,594&6,033\\
\hline
\end{tabular}
\caption{
Corpus statistics showing the counts of token and utterance types across recordings with Arabic-speaking (Rec-Ar) and English-speaking (Rec-En) Interlocutors. The reported total is the summation of both in addition to the pilot recording.
}
\label{table:corpus_stats}
\end{table}

\section{Code-switching Analysis}\label{sec:cs_stats}

\begin{table}[t]
\centering
\setlength{\tabcolsep}{4pt}
\begin{tabular}{|l|r|r|}
    \hline
      \multicolumn{1}{|c|}{\textbf{Metric}}&\multicolumn{1}{c|}{\textbf{Average}}&\multicolumn{1}{c|}{\textbf{SD}}\\
      \hline\hline
      Code-Mixing Index (CMI) &0.20&0.13\\
      Switch Point Fraction (SPF)&0.20&0.14\\
      Percentage of English words&44.0\%&32.6\%\\
      \hline
\end{tabular}
\caption{Arabic-English code-switching statistics, reporting CMI, SPF and percentage of English words, calculated as the mean and standard deviation over code-switched utterances.}
\label{table:cs_stats}
\end{table}
In this section, we present analyses for Arabic-English and MSA-dialectal Arabic code-switching. Statistics on the former code-switching type is presented in Table~\ref{table:cs_stats}.
Table \ref{table:examples_LID} demonstrates examples, covering different dialectness level annotations and showing both types of code-switching.

\subsection{Arabic-English Code-switching}
As reported in the utterance-level analysis presented in Table \ref{table:corpus_stats}, Arabic-English code-switched utterances constitute 19.0\% of all non-annotation-only utterances. 
In order to measure the amount of Arabic-English code-switching, we use three metrics: Code-mixing Index (CMI) \citep{GD16}, Switch Point Fraction (SPF) \citep{PBC+18}, and the percentage of English words. Afterwards, we analyze the morphological code-switching constructs.
\paragraph{Code-Mixing Index}
CMI measures the level of mixing between languages, and is defined on the utterance-level as follows:
\[C(x)=\frac{\frac{1}{2}*(N(x)-max_{L_i\in \textbf{L}}\{t_{L_i}\}(x))+\frac{1}{2}P(x)}{N(x)} \]

where $N$ is the number of language-dependent tokens in utterance $x$; $L_i \in \textbf{L}$ the set of languages in the corpus; $max\{t_{L_i}\}$ represents the number of tokens in the dominating language in $x$, with $1 \leq max\{t_{L_i}\}$$ \leq N$; and $P$ is the number of code alternation points in $x$; $0 \leq P < N$. The corpus-level CMI is calculated as the average of utterance-level CMI values. 
Among the code-switched utterances, the value is 0.20 with standard deviation of 0.13.

\paragraph{Switch Point Fraction}
SPF is calculated as the number of switch points over the number of word boundaries in an utterance. The corpus-level SPF is calculated as the average SPF values over utterances. Among the code-switched utterances, the SPF value is 0.20 with standard deviation of 0.14.
\paragraph{Percentage of English}
Given that the CMI metric does not distinguish between the primary and secondary languages, we report the percentage of English words to get a better understanding on the amount of English usage. Among the code-switched utterances, the 
average percentage of English words over utterances is 44.0\% with standard deviation of 32.6\%.

\begin{table*}[t]
\centering
\begin{small}
  
\begin{tabular}{|l|r|}
\hline
\textbf{Level}&\multicolumn{1}{c|}{\textbf{Example}}\\\hline\hline
\textbf{L0}&\<كيف يمكن أن نشجع عملهم كفريق ولكن في نفس الوقت نحافظ على التخصص؟ >\\
&\multicolumn{1}{l|}{How can we encourage their work as a team, but at the same time maintain specialization?}\\\hline
\textbf{L1}&
\<
\textbf{حددتوها}
؟>
\textbf{\texttt{-{}-}}\<\textbf{ت}>
\<\%أم أي مخاطر أخرى من وجهة نظركم>\\
&\multicolumn{1}{l|}{\%mm any other risks from your point of view \textbf{y-- you specified them}?}\\\hline
\textbf{L2}&\<شكرا، شكرا لك على الفرصة الحلوة \textbf{هذي}.>\\
&\multicolumn{1}{l|}{Thanks, thank you for \textbf{this} nice opportunity.}\\\hline
\textbf{L3}& \<حاليا.> \textit{semester}\<\textbf{طيب} أنا \textbf{باتكلم} عن مادة آخذها أنا في هذا ال+>\\
&\multicolumn{1}{l|}{\textbf{Ok, I am talking} about a course that I am in this \textit{semester} currently.}\\\hline
\textbf{L4}& \<مثلا. \textbf{صح ما بنبيع حق العالم بس} يعني نعرض منتجاتنا.>
\textit{social media}\<\textbf{ونقدر نفتح}
حسابات في ال+>\\
&\multicolumn{1}{p{14cm}|}{\textbf{We can open} accounts on \textit{social media}, for example. \textbf{It's true we will not sell to} the world, \textbf{but} I mean, we offer our products.}\\\hline
\end{tabular}
\caption{Examples of utterances receiving \textit{L0}-\textit{L4} dialectness level annotations. Dialectal words are bolded.}
\label{table:examples_LID}
  
\end{small}
\end{table*}

\paragraph{Morphological Code-switching}
Among the 353 MCS constructs in the corpus, 
we report the attachment of the Arabic definite article \<ال>~\textit{Al}~`the' as the dominating construct, where 78.5\% of the MCS constructs are \textit{Al}+English word. This is in-line with the figures reported for Egyptian Arabic-English code-switching in \citet{HDL+22}, where this construct had a share of around 74\%. Other MCS constructs include 
the attachement of Arabic conjunction proclitics (7.9\%), prepositional proclitics (2.0\%), and feminine plural suffix (1.4\%). The remaining 10.2\% include the use of the definite article in combination with the other mentioned proclitics, plural suffix, as well as cliticized demonstrative pronoun. While MCS can also involve verbal inflection, we do not observe such constructs in our corpus.

\subsection{MSA-Dialectal Code-switching}

With regards to the dialectness level annotation task outlined in 
Section~\ref{sec:LID_annotation}, a total of 1,158 utterances were annotated. 
We report that 40.7\% of the utterances are annotated as \textit{L0}, 17.4\% as \textit{L1}, 11.5\% as \textit{L2}, 14.2\% as \textit{L3}, and 16.2\% as \textit{L4}. 

\section{Morphological Annotation}\label{sec:morhological_analysis}
We automatically annotate the corpus for tokenization, POS tags, and lemmas following \citet{habash2022zaebuc}. We discuss the annotation process as well as analyses and observations.

\subsection{Annotation Guidelines}
For \textbf{English words}, we use Stanza \citep{qi2020stanza} to obtain lemmas and Universal Dependency (UD) POS tags \citep{Nivre:2017:universal}. English tokenization is minimally intrusive; one exception is
contractions, which Stanza tokenizes, e.g., \textit{I'll} is separated to \textit{I+'ll}. 
%
For \textbf{Arabic words}, we first morphologically disambiguate using 
CAMeL Tools BERT-based model \cite{obeid:2020:camel,inoue:2021:morphosyntactic}, which produces lemmas and a wide range of features. We utilize the MSA model for Arabic-speaking Interlocutors, as that is their intended language. For Students' utterances, we use the Gulf Arabic model, and back off to the MSA model in case of missing analyses.
For tokenization, we follow the ATB tokenization, where all clitics are tokenized except for the definite article \citep{Maamouri:2004:patb,Habash:2010:introduction}.
For POS tags, we use UD \cite{Taji:2017:universal}. 
In the case of MSA, these features are readily provided by CAMeL Tools. But for Gulf Arabic, we convert the Buckwalter tag features to get  tokenization and UD POS tags using the mapping provided by \citet{Taji:2017:universal}.
Finally, we use the diacritized Arabic lemmas produced by CAMeL Tools.

As part of the pre-processing step for this annotation task, the transcriptions were automatically white-space-and-punctuation tokenized, except for the following annotation tokens, which were left untokenized: \textit{GR-background}, \textit{CR-nonSpeech}, \textit{CR-interjections}, \textit{CR-partial}, \textit{CR-flow}, \textit{CSR-MCS}, and \textit{ENR-acronyms}. 





\subsection{Statistics and Observations}
\begin{figure}[t]
\centering
 \includegraphics[width=\columnwidth]{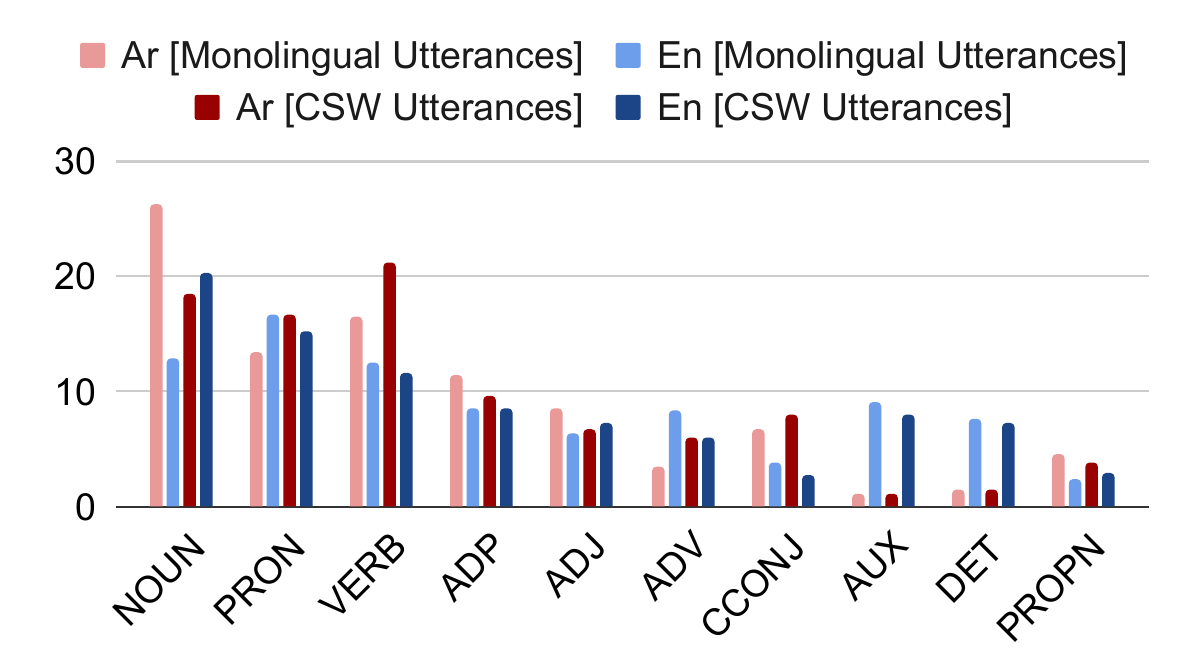}
   \caption{POS distribution for Arabic and English words in monolingual and code-switched (CSW) utterances.}
  \label{fig:pos_distribution}
\end{figure}
\paragraph{Part-of-Speech} In Figure \ref{fig:pos_distribution}, we present the distribution of top-occurring POS tags for Arabic and English words across monolingual and code-switched utterances.\footnote{We report the distribution for Arabic and English words only, excluding partial words, punctuation, annotations, and morphologically code-switched words.} In the context of monolingual utterances, we report that Arabic has a higher usage of NOUN and CCONJ over English, while DET and AUX are more common in English than Arabic. These observations were previously reported and justified in \citet{habash2022zaebuc}. However, unlike 
them,
we report Arabic having a higher frequency of VERBs over English. 
In the context of code-switching, we see a significant increase in frequency for English NOUNs over their occurrence in monolingual utterances, showing +56.9\% relative increase. The prevalence of NOUNs in embedded code-switched words is expected 
as NOUNs are widely-used in borrowing. However, we note that in our corpus the percentage of NOUNs in code-switched utterances (20\%) is lower than that previously reported in \citet{bali2014borrowing} and \citet{hamed2018collection} (57-66\%). This could reflect that the code-switching in the corpus involves more complex intra-sentential code-switching than borrowing of NOUNs. It can also be due to a significant amount of code-switched utterances being dominated by English, thus the POS distribution of the English words being closer to that of monolingual English utterances. This is shown in Table~\ref{table:corpus_stats}, where we report that code-switched utterances having 1-20\% English words only constitute 35.9\% of the code-switched utterances, while the rest are well distributed across the other ranges, with 33.9\% of the code-switched utterances being dominated (61-99\%) by English.

\begin{figure}[t]
\centering
 \includegraphics[width=0.95\columnwidth]{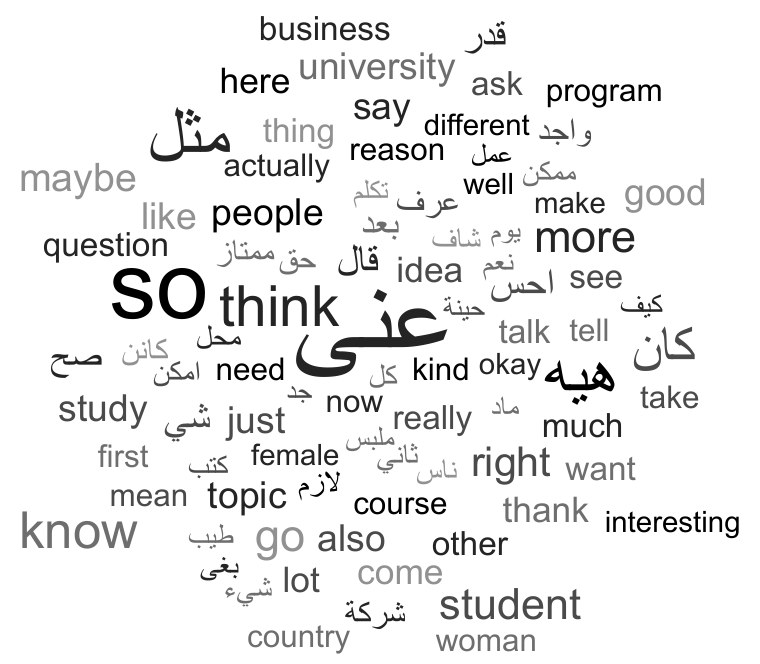}
   \caption{Lemma cloud for the top-occurring 100 noun, verb, adjective, or adverb lemmas. 
   }
  \label{fig:lemma_cloud}
\end{figure}

\paragraph{Tokenization} Under the utilized tokenization schemes, we find that on average, Arabic and English words have 1.23 and 1.03 morphemes, respectively. For English, 96.6\% of the words do not get tokenized, 3.4\% have 2 morphemes, and 0.03\% have 3 morphemes. 
The figures for Arabic words are higher, where the percentages of words consisting of 1, 2, 3, and 4 morphemes are 78.7\%, 19.8\%, 1.5\%, and 0.02\%. It is to be noted that the figures for Arabic do not fully reflect its morphological richness, 
as affixes are not tokenized.

\paragraph{Lemmatization}
In Figure \ref{fig:lemma_cloud}, we show the lemma cloud for the 100 top-occurring noun, verb, adjective, or adverb lemmas in the corpus. 
The top-occurring lemmas constitute words that are normally frequently present in speech in addition to words that are specific to the discussed topics. The former case includes lemmas such as \<عنى> \textit{{\AYN}n{\AMAQSURA}} `mean', \<مثل> \textit{m{\THA}l} `like', \<هيه> \textit{hyh} `yes', \<كان> \textit{kAn} `be', \textit{so}, \textit{have}, \textit{think}, and \textit{know}. The latter includes lemmas such as 
\<شركة> \textit{{\SHIN}rk{\TAMARBUTA}} `company', 
\<محل> \textit{mHl} `shop', 
\<ناس> \textit{nAs} `people', 
\<عمل> \textit{{\AYN}ml} `work', 
\textit{student}, \textit{study}, \textit{university}, and \textit{business}.

\section{Conclusion and Future Work}\label{sec:conclusion}
In this work, we extend the currently available speech corpora with a bilingual, multidialectal corpus, containing (accented) English and Modern Standard Arabic, as well as Gulf and Egyptian Arabic dialects. The recordings are collected through Zoom meetings and are manually transcribed. 
We develop our transcription guidelines to handle  challenges introduced by conversational speech, code-switching, and unstandardized orthography. We provide an analysis on the code-switching involved in the corpus. We also automatically annotate the corpus for POS tags, tokenization and lemmatization, and plan on extending the corpus with manual revisions. Also, for the subset of the corpus containing code-switching between Arabic variants, we provide dialectness level annotations. 

Looking ahead, we plan to manually annotate the corpus for morphological features and syntactic representations. We also plan to use it to benchmark a range of NLP tasks from ASR to morphological disambiguation and syntactic parsing.

\section*{Ethics Statement}
We conducted this research with full approval from Zayed University  Institutional Review Board. All annotators and participants were compensated fairly according to campus payment policies. 
Participants were informed that the meetings are being recorded, and that the purpose of the study is collection of natural speech. 
Participants gave explicit consent for the public release of audio files and transcriptions. We've taken rigorous steps to anonymize any sensitive or identifying information.

\section*{Acknowledgments}
ZAEBUC was funded by a 
Zayed University Research Incentive
Fund (RIF~\#R19068).  We thank
Dr. Michael Bowles for his project management help.

\section{Bibliographical References}\label{sec:reference}
\bibliographystyle{lrec-coling2024-natbib}
\bibliography{custom}
\section{Language Resource References}
\label{lr:ref}
\bibliographystylelanguageresource{lrec-coling2024-natbib}
\bibliographylanguageresource{languageresource}

\end{document}